\pdfoutput=1
\documentclass[11pt]{article}

\usepackage{ACL2023}
\usepackage{times}
\usepackage{latexsym}
\usepackage{amsmath}
\usepackage{amssymb}
\usepackage{booktabs}
\usepackage[T1]{fontenc}
\usepackage[utf8]{inputenc}

\usepackage{microtype}
\usepackage{adjustbox}
\usepackage{inconsolata}

\usepackage{enumitem}
\usepackage{amsmath}
\usepackage{amssymb}

\usepackage{multirow}

\title{PEFT-Ref: A Modular Reference Architecture and Typology\\for Parameter-Efficient Finetuning Techniques}
\author{Mohammed Sabry \\
  ADAPT/DCU, Dublin, Ireland \\
  \texttt{mohammed.sabry@adaptcentre.ie}\And
  Anya Belz \\
  ADAPT/DCU, Dublin, Ireland\\
  \texttt{anya.belz@adaptcentre.ie} \\}

\begin{document}
\maketitle
\begin{abstract} 
Recent parameter-efficient finetuning (PEFT) techniques aim to improve over the considerable cost of fully finetuning large pretrained language models (PLM). As different PEFT techniques proliferate, it is becoming difficult to compare them, in particular in terms of (i) the structure and functionality they add to the PLM, (ii) the different types and degrees of efficiency improvements achieved, (iii) performance at different downstream tasks, and (iv) how differences in structure and functionality relate to efficiency and task performance.
To facilitate such comparisons, this paper presents a reference architecture which standardises aspects shared by different PEFT techniques, while isolating differences to specific locations and interactions with the standard components. Through this process of standardising and isolating differences, a modular view of PEFT techniques emerges, supporting not only direct comparison of different techniques and their efficiency and task performance, but also systematic exploration of reusability and composability of the different types of finetuned modules. 
%We apply the framework to current PEFT techniques to provide a detailed comparison of their structural and functional properties, efficiency and task performance, illustrating 
We demonstrate how the reference architecture can be applied to understand properties and relative advantages of PEFT techniques, hence to inform selection of techniques for specific tasks, and design choices for new PEFT techniques. 
\end{abstract}

\section{Introduction}

Over the past few years, there has been a significant increase in the size of pretrained language models (PLMs) such as GPT3 \cite{https://doi.org/10.48550/arxiv.2005.14165}, OPT \cite{ https://doi.org/10.48550/arxiv.2205.01068}, BLOOM \cite{https://doi.org/10.48550/arxiv.2211.05100}, and PaLM \cite{https://doi.org/10.48550/arxiv.2204.02311}, which have billions of parameters. This increase in size has been accompanied by a commensurate increase in the cost of training and deploying large PLMs, with substantial financial and environmental implications. Reusing PLMs via adaptation to downstream tasks, rather than training new language models for new tasks, mitigates this cost significantly. However, full finetuning, the default task adaptation approach, is still very costly as it retrains, and subsequently stores, the entire model. 
%These models are often provided as a service, making it challenging to increase their reusability and mitigate high training costs. To adapt these models, transfer learning via finetuning is the prevalent framework. However, full finetuning may be impractical to serve different downstream tasks and users.

Parameter-efficient finetuning (PEFT) techniques improve this cost by (re)training a much smaller set of parameters. Heuristic approaches modify a specific subset of the model's existing parameters, e.g.\ \citet{lee2019elsa} finetune the last quarter of the layers in BERT and RoBERTa, and \citet{zaken2022bitfit} finetune just the bias terms of the model. Other PEFT techniques such as Adapters \cite{pmlr-v97-houlsby19a}, prefix tuning \cite{li-liang-2021-prefix}, prompt tuning \cite{lester-etal-2021-power}, and LoRA \cite{DBLP:journals/corr/abs-2106-09685}, instead freeze all of the PLM parameters, and add and train a small set of new parameters in conjunction with the latter. Several studies \cite{Ding2023, chen-etal-2022-revisiting, he2022towards, mao-etal-2022-unipelt} have %
%focused on exploring 
found such parameter-adding PEFT techniques highly effective in real-world tasks. It is this group of PEFT methods  that is our focus in this paper. 

As an increasing number of PEFT techniques are reported, it is becoming harder to compare them in terms of efficiency improvements and performance at different tasks, in particular which aspects of their structure and functionality are linked to better efficiency and performance. To address this we propose the PEFT-Ref framework  
%aim to emphasise the modular perspective of PEFT techniques, and 
consisting of a modular reference architecture and typology which provide a standardised way of characterising PEFT techniques in terms of their structural and functional properties. %, thus facilitating comparison in terms of efficiency and task performance. %We demonstrate the practical use of PEFT-Ref for recommending techniques for downstream tasks and guiding the creation of new techniques.
In this paper, we present the reference architecture (Section~\ref{sec:peft-ref}), and use it to create a typology of seven leading PEFT techniques (Section~\ref{section:peft-techniques}), and to compare the techniques in terms of efficiency and performance (Section~\ref{sec:peft_comparison}). We illustrate how this in turn can be used to inform design choices and technique selection for specific tasks (Sections~\ref{section: peft_selc_dev}), and finish with a review of related work 
%on finetuning and modularity 
and some conclusions (Section~\ref{section:related_work} and~\ref{sec:conclusions}).
%The paper is structured as follows: Section~\ref{sec:peft-ref} presents PEFT-Ref 
%%, a framework for characterising and comparing PEFT techniques, 
%in detail. Section~\ref{section:peft-techniques} uses PEFT-Ref to characterise seven leading PEFT techniques. Section~\ref{sec:peft_comparison} compares the techniques in terms of efficiency and performance, and
%%using PEFT-Ref. 
%Section~\ref{section: peft_selc_dev} illustrates using PEFT-Ref to inform design choices and technique selection for specific tasks. Section~\ref{section:related_work} reviews related work on finetuning and modularity and situates our work within these areas.

%\section{PEFT-Ref: A Reference Architecture for PEFT Techniques}\label{sec:peft-ref}
\section{The PEFT-Ref Framework}\label{sec:peft-ref}

In this section, we present the PEFT-Ref in diagrammatic form (Section~\ref{sec:ref-arch}), and  in terms of the typological properties it defines (Section~\ref{sec:typology}). In combination, reference architecture and properties are intended to fully capture differences and similarities between different PEFT techniques, as a basis for understanding the causes for their relative strengths and weaknesses, and informing technique selection and development.

%We use PEFT-Ref to characterise  seven leading PEFT techniques in terms of their structural properties, and how they individually interact with the standard architecture, in  Section~\ref{section:peft-techniques}.

\subsection{Modular PEFT reference architecture}\label{sec:ref-arch}

%%% 1. the diagram
%%% 2. detailed description of the diagram

\begin{figure*}[t]
    \centering
    \includegraphics[width=1.05\textwidth]{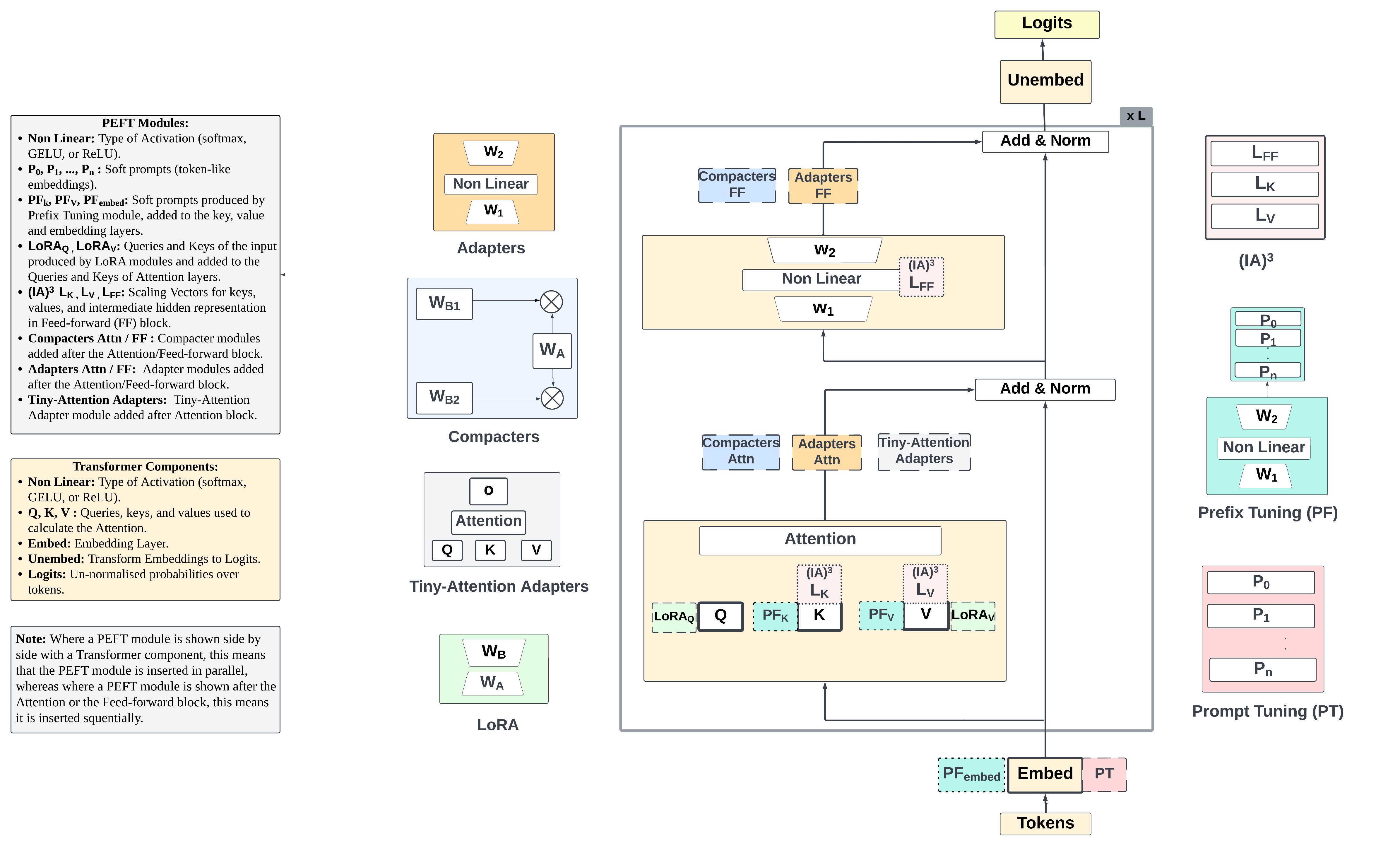}
    \caption{Modular PEFT reference architecture showing PLM components (central box), different types of PEFT modules (left and right of centre), and insertion slots of PEFT modules (dashed boxes in PLM box) and interactions between PEFT modules and PLM components (see also legend on left).}
    \label{fig:peft}
\end{figure*}

Figure~\ref{fig:peft} displays the PEFT reference architecture in diagrammatic form, showing how the different types of modules created and trained by different PEFT techniques slot into and interact with the standard Transformer architecture. Most of the properties defined in the 
%typology (Section~\ref{sec:typology}) 
next section are also depicted in the diagram (see the Figure~\ref{fig:peft} legend).

In the diagram, the \textit{L $\times$} repeated layers are shown inside the grey box, with the residual flow to the right. The components of the standard Transformer PLM are shown in black (non-dashed) boxes. In the embedding, attention and feed-forward layers, and immediately following the attention and feed-forward layers, we show where and how different PEFT techniques insert their modules, indicated by dashed boxes. Note that these are not normally combined, i.e.\ only one type of PEFT module is normally inserted. 

%Section~\ref{section:peft-techniques} describes in more detail how each of the seven types of PEFT module depicted in the diagram is structured internally and collaborates with the PLM.

\subsection{Modular properties of PEFT modules}\label{sec:typology}\label{section: peft-ref}

%%% 3. list of modular properties with direct reference to the diagram

The PEFT-Ref typology comprises the following modular structural and functional properties. We adapt some of the property \textit{names} from the general modular computing literature, some from recent work on modularity in neural networks, and some are new, as indicated. The range of property \textit{values} is specific to PEFT-Ref in all cases. Table~\ref{tab:modularity structural properties} lists these properties and their specific values for seven leading PEFT techniques. 

\begin{enumerate}[itemsep=0pt]
    
\item \textbf{Intra-connectivity} \cite{Clune2013,10.3389/fnins.2010.00200} -- \textit{dense} or \textit{sparse}: Neuron connectivity within the PEFT module's layers. Denser intra-connectivity indicates higher modularity. All current PEFT techniques, except (IA)\textsuperscript{3}, are densely intra-connected. In Table~\ref{tab:modularity structural properties}, we additionally show the specific type of densely connected component: \textit{embedding layer, non-linear MLP, linear MLP, self-attention}; (IA)\textsuperscript{3} inserts a standalone vector parameter that is neither dense nor sparse.

\item \textbf{Inter-connectivity} \cite{https://doi.org/10.48550/arxiv.2106.02626, 10.3389/fnins.2010.00200} -- \textit{fixed:dense, fixed:sparse} or \textit{dynamic}:  How PEFT modules are connected to the PLM architecture. Sparser inter-connectivity indicates higher modularity. All current PEFT techniques except tiny-attention adapters \cite{zhao-etal-2022-tiny} have fixed/dense interconnectivity.

\item \textbf{Parameters adapted} \cite{Ding2023} -- \textit{addition} or \textit{reparameterisation}: All PEFT techniques alter the model parameters, either by adding them or reparameterising existing components in the PLM architecture. %This accounts for two observations: the performance of blindly adding parameters for adaptation versus reparameterising specific components. Reparameterised parameters inserted in parallel can be seamlessly composed with the language model, avoiding inference latency\footnote{In Appendix \ref{sec:complexity_analysis}, we discussed the implications of inference latency for PEFT techniques.}.

\item \textbf{Parameter Sharing/Tying} -- \textit{shared}, \textit{tied}, or \textit{none}: In parameter sharing two sets of parameters are forced to be the same; in tying, two sets of parameters are kept close to each other. Parameter sharing/tying has several advantages in regularisation and inductive biases \cite{pmlr-v151-yeh22b}, including better performance and stability with fewer parameters. Among current PEFT techniques, only Compacters \cite{NEURIPS2021_081be9fd} share parameters, in their reparameterised layers. %(Compacters parameterise Adapter layers, resulting in 4 small layers, but instead of maintaining 4 layers, they share a layer, resulting in 3 layers as shown in Figure~\ref{fig:peft}, with layer $W_A$ shared between layers $W_{B1}$ and $W_{B2}$).

\item \textbf{Input type} \cite{pfeiffer2023modular, AUDA1999} -- \textit{hidden}, \textit{data} or \textit{weights}: The type of input PEFT modules receive: %. Current PEFT techniques take three types of input: 
(i) hidden representations received from a Transformer layer block, (ii) the data before it goes into the block, or (iii) a newly initialised weight matrix, in the case of PEFT techniques that add and optimise a weight matrix (Prompt Tuning, Prefix Tuning, and (IA)\textsuperscript{3}). 
    
\item \textbf{Insertion Form} \cite{pfeiffer2023modular, AUDA1999} -- \textit{sequential} or \textit{parallel}: Whether the finetuned module is inserted into the PLM sequentially or in parallel. Most techniques that insert modules sequentially receive the output of the Transformer layer block they collaborate with.
    
\item \textbf{\#Insertions} -- \textit{n layers} or \textit{all layers}: How many instances of a PEFT module are inserted into the PLM. All current PEFT techniques except prompt tuning (which inserts only into the embedding layer), insert modules into all of the $L \times$ repeated Transformer layers. Prefix tuning additionally adds parameters to the embedding layer.
    
\item \textbf{Integration Form} \cite{AUDA1999} -- \textit{concatenation, scaled addition},  \textit{direct addition}, \textit{gated addition}, or \textit{rescaling}: How PEFT module outputs are incorporated into the PLM.
    
\item \textbf{Workspace} -- \textit{attention layer}, \textit{FFN layer} or \textit{embedding layer}: In cognitive science, workspace is a limited-bandwidth communication channel in which different modules exchange information \cite{Baars1988-BAAACT}. In AI, \citet{goyal2022coordination} use a shared workspace model to describe systematic information exchange between specialist regions in a neural network. In our context, most PEFT techniques use attention layers and/or fully connected layers in the PLM as their workspace. Table~\ref{tab:modularity structural properties} additionally indicates, where appropriate, the specific locus of interaction within the workspace -- \textit{queries/values}, \textit{keys/values}, \textit{(FFN) intermediate representation}.

\end{enumerate}

\begin{table*}
\centering
\begin{adjustbox}{width=\textwidth,center=\textwidth}
\begin{tabular}{lccccccccc}
\toprule
\textbf{PEFT} & \textbf{Intra-} & \textbf{Inter-} & \textbf{Parameters} & \textbf{Parameter} & \multirow{2}{*}{\textbf{Input type}}    & \textbf{Insertion} & \multirow{2}{*}{\textbf{\#Insertions}} &\textbf{Integration} & \multirow{2}{*}{\textbf{Workspace}} \\
\textbf{technique} & \textbf{connectivity} & \textbf{connectivity} & \textbf{adapted} & \textbf{Sharing} & \textbf{}    & \textbf{form} & \textbf{} &\textbf{form} & \textbf{} \\
\midrule
Prompt tuning   & dense:embedding              & fixed:dense                       & Addition    & none            & Weights                   & Parallel    & 1 layer           & concatenation             & embedding layer         \\
\multirow{2}{*}{Prefix tuning}   & \multirow{2}{*}{dense:non-linear MLP}             & \multirow{2}{*}{fixed:dense}                       & \multirow{2}{*}{Addition}  & \multirow{2}{*}{none}             & \multirow{2}{*}{Weights}                   & \multirow{2}{*}{Parallel}   & \multirow{2}{*}{all layers}             & \multirow{2}{*}{gated addition}            & embedding layer; \\
&&&&&&&&& att.\ layer:keys/values \\
LoRA            & dense:linear MLP                  & fixed:dense                       & Reparametr.\   & none    & Data                   & Parallel  & all layers               & scaled addition            & att.\ layer:queries/values      \\
Adapters        & dense:non-linear MLP              & fixed:dense                       & Addition  & none               & Hidden & Sequential    & all layers           & direct addition           & FFN layer; attention layer           \\
Tiny-Att.\ Ad.\ & dense:self-attention             & dynamic                       & Addition  & none               & Hidden  & Sequential   & all layers            & direct addition           & attention layer           \\

Compacters   & dense:non-linear MLP            & fixed:dense                       & Addition  & shared             & Hidden                    & Sequential  & all layers               & direct addition            & FFN layer; attention layer \\
\multirow{2}{*}{(IA)\textsuperscript{3}} & \multirow{2}{*}{none:parameter vector} & \multirow{2}{*}{fixed:dense} & \multirow{2}{*}{Addition} & \multirow{2}{*}{none} & \multirow{2}{*}{Weights} & \multirow{2}{*}{Sequential} & \multirow{2}{*}{all layers}  & \multirow{2}{*}{rescaling} & FFN layer:intermed.\ repres.; \\
&  &  &  &  &  &  &   &  &  attention layer:keys/values  \\
\bottomrule
\end{tabular}

\end{adjustbox}
\caption{Structural properties of PEFT modules created by seven PEFT techniques (for descriptions of techniques see Section~\ref{section:peft-techniques}; for definitions of properties see Section~\ref{section: peft-ref}).}
\label{tab:modularity structural properties}
\end{table*}

\section{Characterisation of PEFT Techniques with PEFT-Ref}\label{section:peft-techniques}

%%% 4. description of individual PEFT techniques:\\
%%% 4a. how they interface with the standard Transformer components, precisely referring to the diagram\\
%%% 4b. what their modular properties are, precisely referring to the list of modular properties\\

In this section, we characterise seven leading PEFT techniques in terms of PEFT-Ref modular structural properties. %Each technique augments the standard Transformer architecture with additional parameters that are trained during finetuning (while the base model parameters are frozen). Figure~\ref{fig:peft} shows where each of the seven techniques inserts these parameters into the PLM (in some cases in multiple locations), and how they interface with the model architecture and data flow. 
We focus on PEFT techniques that are modular in the sense that they add and train topologically distinct sets of parameters. Such techniques have been shown to be highly effective and are commonly used in downstream and real-world tasks \cite{Ding2023, liu2022fewshot}, in contrast to more heuristic approaches that adapt a specific fixed subset of PLM parameters \cite{lee2019elsa,zaken2022bitfit}. 
%We leave out non-modular techniques that  finetune a subset of the existing model's parameters, e.g.\ only finetuning the last quarter of the layers in BERT and RoBERTa \cite{lee2019elsa}, or only finetuning the bias terms of the model \cite{zaken2022bitfit}.
Table~\ref{tab:modularity structural properties} provides an overview of %the structural properties 
the seven PEFT techniques covered in this section, in terms of the typological properties introduced in Section~\ref{sec:typology}.

\subsection{Prompt Tuning (PT)}

\textit{Module-internal topology and functionality:} Prompt Tuning (PT) \cite{lester-etal-2021-power} generates token-like embeddings 
%(weights) 
using an embedding layer (intra-connectivity), which are then concatenated (integration form) to the input embeddings of the PLM (workspace), as shown in Figure \ref{fig:peft}. The finetuning process customises the token-like embeddings to the task objective.

\textit{Modular properties and collaboration with PLM:} the %The modular properties of PT are summarised in Table~\ref{tab:modularity structural properties}, Row~1. 
PT  only inserts parameters at the embedding layer, concatenating all token-like embeddings with the input embeddings, which results in a fixed-dense inter-connectivity between the PLM and PT.

\subsection{Prefix Tuning (PF)\footnote{Prefix tuning was published prior to prompt tuning, but the two appear to have been developed simultaneously and, coincidentally, the former is an enhanced variation of the latter.}}

\textit{Module-internal topology and functionality:} In contrast to Prompt Tuning, which generates token-like embeddings using only the embedding layer, 
%as intra-connectivity, 
\citet{li-liang-2021-prefix} propose using two linear layers with a Softmax activation in-between (Figure~\ref{fig:peft}).

\textit{Modular properties and collaboration with PLM:} %Table~\ref{tab:modularity structural properties}, Row~2  summarises the modular properties of PF. 
\citet{li-liang-2021-prefix} furthermore extend the workspace to be the input embeddings and the Attention's keys and values in all Transformer layers. The PF token-like embeddings are concatenated with these matrices ( i.e. integration form).\footnote{\citet{he2022towards} demonstrate that the concatenation in the Attention block can be viewed as a form of gated addition integration; in \textit{$(1-\lambda)h + \lambda\Delta h$} the $h$ represents PLM functionality and $\Delta h$ represents PEFT module functionality.} PF connects all its information to the PLM (i.e.\ its inter-connectivity is fixed:dense).

\subsection{LoRA}

\textit{Module-internal topology and functionality:} LoRA \cite{DBLP:journals/corr/abs-2106-09685} 
%stands for `Low-Rank Adaptation of LLMs'. LoRA aims to 
adapts PLMs using low-rank decomposition matrices. The idea  is that the update of model parameters can be approximated using low-dimensional decomposition. LoRA reparameterises the Attention queries and values weights into low-rank matrices. For each, LoRA uses two small linear projection layers (inter-connectivity) to reparameterise the weights. LoRA receives the same input that the reparameterised weights receive (i.e.\ the insertion form is parallel). 

\textit{Modular properties and collaboration with PLM:} %Table~\ref{tab:modularity structural properties}, Row~3 summarises the modular properties of LoRA. 
LoRA generates queries and values of the input and collaborates (integration form) via scaled addition \textit{($h + \lambda\Delta h)$}  with the Attention's queries and values (workspace) of the input in all Transformer layers. LoRA sends all its information to the workspace (i.e.\ inter-connectivity = fixed:dense).

\subsection{Adapters}

\textit{Module-internal topology and functionality:} Adapters \cite{pmlr-v97-houlsby19a} use a feed-forward layer (intra-connectivity) that bottlenecks information via two linear layers that project the information down and then up, with ReLU activation in-between. Adapters adapt the hidden representations resulting from Attention and FNN blocks (insertion form = sequential).

\textit{Modular properties and collaboration with PLM:} %Table~\ref{tab:modularity structural properties}, Row~4 summarises the modular properties of Adapters. 
Adapters integrate their results with their workspace (Attention and FNN blocks) via direct addition \textit{($h + \Delta h)$}. Although variants of Adapters exist that change internal connectivity or \#insertions, such as AdapterDrop \cite{ruckle-etal-2021-adapterdrop}, Compacters \cite{NEURIPS2021_081be9fd}, and Tiny-Attention Adapters \cite{zhao-etal-2022-tiny}, they all use direct addition for integration.  Adapters send all their information to their workspace (inter-connectivity = fixed:dense).

\subsection{Tiny-Attention Adapters}

\textit{Module-internal topology and functionality:} Tiny-Attention Adapters \cite{zhao-etal-2022-tiny} are a variant of Adapters that change the intra-connectivity to a small Attention layer (Figure \ref{fig:peft}).

\textit{Modular properties and collaboration with PLM:} %Table~\ref{tab:modularity structural properties}, Row~5 summarises the modular properties of Tiny-Attention Adapters. 
Like Adapters, Tiny-Attention Adapters are inserted sequentially, collaborate via direct addition with their workspace, and receive hidden representations as inputs. However, they are inserted  after the Attention block (workspace), 
%in all Transformer layers, 
and send their information to the workspace selectively based on the input (inter-connectivity = dynamic). 

\subsection{Compacters}

\textit{Module-internal topology and functionality:} Compacters \cite{NEURIPS2021_081be9fd} are a variant of Adapters with the following difference. In the vanilla Adapter layer, $W \in \mathbb{R}^{k\times d}$. In contrast, Compacters reparameterise layer $W$ as a sum of Kronecker products, with $k$ and $d$ divisible by a user-defined hyperparameter $n$. Specifically, the sum of $n$ Kronecker products is $W = \sum_{i=1}^{n} A_i \otimes B_i$,  where $A_i \in \mathbb{R}^{n\times n}$ and $B_i \in \mathbb{R}^{\frac{k}{n} \times \frac{d}{n}}$.  
Compacters further improve parameter efficiency by sharing the weights of $A_i$ between the layers of the compacter. %This not only saves parameters but also tailors the idea of fast and slow weights, where fast weights capture layer-specific information and slow weights capture general information across layers. 

\textit{Modular properties and collaboration with PLM:}  %Table~\ref{tab:modularity structural properties}, Row~6 summarises the modular properties of Compacters. 
Compacters have the same properties as Adapters in terms of collaboration with the PLM, insertion form, integration form, and workspace. 

\subsection{(IA)\textsuperscript{3}}

\textit{Module-internal topology and functionality:} An (IA)\textsuperscript{3} \cite{liu2022fewshot} 
%refers to "Infused Adapter by Inhibiting and Amplifying Inner Activations". A 
module comprises three vectors that rescale the Attention (keys, values), and FFN blocks of a Transformer layer (Figure \ref{fig:peft}). During the tuning process, these vectors are initialised to one to ensure that the module does not affect the PLM's functionality before being guided by the task's objective gradient.

\textit{Modular properties and collaboration with PLM:}  %Table~\ref{tab:modularity structural properties}, Row~7 summarises the modular properties of (IA)\textsuperscript{3}.
(IA)\textsuperscript{3} applies learned vector rescaling to its workspace (keys, values, and intermediate FFN) across all Transformer layers.  It is inserted sequentially and sends all its information to its workspace (inter-connectivity = fixed:dense).

\section{Efficiency and Performance Comparisons with PEFT-Ref}\label{sec:peft_comparison}

In this section we use PEFT-Ref as the basis for several different types of comparisons between the seven techniques characterised in the preceding section. In Section~\ref{ssec:comp-effic} we take a closer look at exactly what the efficiency improvements achieved by each PEFT technique are, (i) as compared to full finetuning involving all PLM parameters, and (ii) as compared to the other PEFT techniques.

Then in Section~\ref{ssec:comp-perf} we review what we know so far about the performance of the seven techniques at different benchmark tasks, and link it to their modular properties. 

\subsection{Efficiency improvements}\label{ssec:comp-effic}

%%% 5. comparison of the 7 PEFT techniques in terms of efficiency:

%%% 5a. computational complexity

\subsubsection{Complexity} \label{ssec:complexity}

Table \ref{tab:peft-efficient} provides an overview of PEFT techniques in terms of the time complexity per token of the module(s) they add (column~2), and the number of parameters added per Transformer layer (column~3). Module time complexity (column~2) is controlled by intra-connectivity and input type.\footnote{PEFT techniques with $\mathcal{O}(1)$ time complexity output from input in one step. Except for methods that use another network to produce weights, all PEFT techniques that take weights as input and produce weights as output have $\mathcal{O}(1)$ time complexity in this sense.} Here, we take into account only the time it takes a PEFT technique to produce the output for collaboration with the PLM. In this sense, e.g\ (IA)\textsuperscript{3} has constant time complexity $\mathcal{O}(1)$, as the output is obtained directly from the module after weight initialisation, and is used as  a rescaler for the PLM's activations. We provide further details of our anyalysis of module complexity in Appendix~\ref{sec:complexity_analysis}.

The number of parameters (column~3) is chiefly controlled by  workspace type, \#insertions, and intra-connectivity. For example, (IA)\textsuperscript{3} utilises three vectors to rescale the Attention keys, values, and FFN intermediate representations, giving d$_m$ (the model dimension) for keys and values each, plus 4d$_m$ for the FFN intermediate representation, i.e.\ a total of 6d$_m$.

\begin{table}
\centering
\begin{adjustbox}{width=.48\textwidth}
\begin{tabular}{lcc}
\toprule
PEFT                    & Module  & Number of  parameters \\ 
                    &  complexity &  per Transformer layer \\ \midrule
Prompt Tuning           & $\mathcal{O}(1)$            & $n$ $d$$_m$                                   \\
Prefix Tuning           & $\mathcal{O}(kd)$           & $n$ $d$$_m$  + $d$$_m$\textsuperscript{2} + 2$d$$_h$  $d$$_m$                     \\
LoRA                    & $\mathcal{O}(rd)$           & 2 × (2$d$$_h$  $d$$_m$ )                           \\

Tiny-Attention Adapters & $\mathcal{O}({T})$          & 4 × $d$$_m$                                   \\
Adapters                & $\mathcal{O}(kd)$           & 2 × (2 $d$$_h$  $d$$_m$ )                           \\
Compacters               & $\mathcal{O}(\frac{kd}{N})$          & 2 × (2 ($d$$_h$ + $d$$_m$) )                         \\
(IA)\textsuperscript{3}                   & $\mathcal{O}(1)$            & 6 × $d$$_m$                                    \\ \bottomrule
\end{tabular}
\end{adjustbox}
\caption{Efficiency of the seven PEFT techniques surveyed; d$_m$ = model dimension, d$_h$ = PEFT module dimension, $n$ = number of tokens for prompt and prefix tuning;  $k, r, d $ = input/output dimension of PEFT module, where for LoRA $r$ is the rank, and for Adapters $k$ is the bottleneck dimension. $d$ = $d_m$. $T$ = \#Input embeddings. $N$ = Reduction dimension in Kronecker-products.}
\label{tab:peft-efficient}
\end{table}

%%% 5b. efficiency improvements relative to full finetuning and to each other (in-training efficiency)

\subsubsection{In-training efficiency}

Parameter efficiency does not necessarily translate into learning efficiency. \citet{Ding2023} examined the convergence of PEFT techniques such as LoRA, Adapters, Prefix Tuning, and Prompt Tuning against full finetuning. The results showed that full finetuning converged the fastest, followed by Adapters/LoRA, and then Prefix Tuning, while Prompt Tuning had the slowest convergence rate. As PLMs grow in size, the convergence of PEFT techniques becomes faster. \citeauthor{Ding2023}'s results also indicate that convergence is more sensitive to structure than to number of parameters. 

PEFT-Ref explicitly accounts for the structural properties that control convergence rate, including intra/inter-connectivity, \#insertions, output production (input type and insertion form), and parameter sharing. For instance,  slow convergence in Prompt Tuning can be attributed to instability caused by the output (token-like embeddings) being optimised directly, while Prefix Tuning is sensitive to reparameterisation choices (intra-connectivity) that produce this output.  Additionally, some PEFT techniques can have similar or better convergence rates than full finetuning depending on task complexity.

\citet{chen-etal-2022-revisiting} examined the stability of performance across different random seeds for several PEFT techniques including Adapters, LoRA, and Prefix Tuning, following a similar study on the stability of full finetuning \cite{DBLP:journals/corr/abs-2002-06305}. The authors found that these PEFT techniques, like full finetuning, are susceptible to performance fluctuations resulting from weight initialisation and data ordering. Furthermore, the authors investigated the impact of controlling the number of parameters in these techniques on their stability. They observed that reducing the number of parameters in PEFT techniques can increase their stability. As a result, they recommend exercising caution when selecting the reduction factors in Adapters, the rank in LoRA, and the prompt length in Prefix Tuning, and setting them in a low range. In Appendix \ref{sec:complexity_analysis}, we look in more detail at forward \& backward training passes efficiency within the context of module complexity as per Section~\ref{ssec:complexity}.

%%% 5c. storage efficiency and in-application efficiency

\subsubsection{Storage and in-application efficiency}

The last column in Table \ref{tab:peft-efficient} shows the number of parameters added per transformer layer, and the variables that control it, for each of the seven PEFT techniques. 
%For more details on storage size and what controls it in different PEFT techniques, refer to the number of PEFT parameters in . 
By saving only the task-specific post-finetuning PEFT modules\footnote{All PEFT techniques save their tunable parameters, the exception is Prefix Tuning, which only saves the final token-like embeddings and discards the network  that produced them.} instead of the entire model as would be required in full finetuning, storage size can be drastically reduced from gigabytes to a few megabytes. This storage efficiency makes it possible to serve multiple users and applications using a single standalone PLM in conjunction with multiple different task-specific PEFT modules. 

The structural properties defined in PEFT-Ref (e.g., insertions, input type, adapted parameters, workspace, parameter sharing) directly control efficiency in this sense, thus facilitating insights for potential improvements. In Appendix~\ref{sec:complexity_analysis}, we look in more detail at the inference latency for in-application efficiency within the context of module complexity as per Section~\ref{ssec:complexity}.

\subsection{Task performance}\label{ssec:comp-perf}

%%% 6. comparison of the 7 PEFT techniques in terms of what's known about their task performance, directly relating it to the modular properties and efficiencies
In Table \ref{tab:peft-performance} in Appendix \ref{sec:performance}, we have documented the performance of various PEFT techniques across different tasks based on previous research.

Among the techniques we examined, LoRA stands out as the top performer in several tasks, either as the first or the second-best option. LoRA works in collaboration with the PLM to improve critical components, particularly the queries matrices. It is the only PEFT technique that collaborates with this component. 

Adapters, and their variants, also exhibit excellent performance scores, and it appears that the reparameterisation and parameter-sharing properties in Compacter enhance their effectiveness.\footnote{Adapters and Compacters differ in parameter sharing and reparameterisation, with Compacters being more performant than Adapters. These properties are responsible for their performance, as shown in the performance Table \ref{tab:peft-performance}.} Finally, we observe that (IA)\textsuperscript{3}  performs better in commonsense reasoning tasks compared to LoRA and Adapters which could be attributable to the former using rescaling as its integration form, and the latter using addition. 

LoRA, Adapters, and Compacter use either just  attention layers, or attention layers and FFN layers, as their workspace and our analysis indicates that PEFT techniques that use feed-forward and/or Attention blocks as their workspace are associated with higher performance scores.

\section{Using PEFT-Ref to Guide Technique Selection and Development}\label{section: peft_selc_dev}

In this section, we start from (i) the information that PEFT-Ref provides about PEFT techniques, and (ii) their performance at different downstream tasks, to draw broad conclusions about the suitability of each technique for different task types (Section~\ref{ssec:tech-selec}).

Then we take this one step further and surmise how PEFT techniques (Section~\ref{ssec:tech-devel}) can potentially be developed further, or even combined, to improve their stability, convergence speed, and/or task performance.

\subsection{PEFT technique selection}\label{ssec:tech-selec}

Prompt Tuning is a suitable technique for task like Named Entity Recognition, because it works on the embedding layer, which already has enough contextual information to solve this task, after propagating through the frozen language model layers. This means that conditioning the task on the embeddings alone is sufficient. Additionally, Prompt Tuning has a layer complexity of $\mathcal{O}(1)$ and a low number of parameters, making it an efficient option that can achieve good performance even with a small computation budget.

LoRA can be a suitable choice for Question Answering tasks as it operates on the attention queries and values workspaces which enables the model to identify relevant relationships between words and phrases in the question and the answer (LoRA's performance in multiple-choice QA tasks Table~\ref{tab:peft-performance} supports this). Additionally, the tunable scaling integration form can assist the model to better utilise important information to solve the task. Tiny-Attention Adapters can provide additional attention and potentially improve upon the hidden representation output after the Transformer attention block as they are inserted sequentially after it.

Data-to-Text and Summarisation tasks can benefit from using either LoRA or Prefix Tuning. Previous research \cite{li-liang-2021-prefix, liu2022fewshot, xu-etal-2022-evaluating, Ding2023} has shown that these techniques provide comparable performance, but the choice between them depends on the available computation budget. LoRA has fewer parameters and better layer complexity compared to Prefix Tuning, which makes it a more efficient option, and their performance on these tasks can be explained by their properties in PEFT-Ref. Recent work \cite{xu-etal-2022-evaluating} evaluated Adapters for generation tasks and found that although they have good performance, they have worse faithfulness scores than full finetuning and Prefix Tuning. %, which is the best one among them. 
To explain these results in light of PEFT-Ref, it can be noted that Adapters use the feed-forward block in addition to the attention block as their workspace. However, \citet{zhang-etal-2022-moefication} found that the feed-forward block contains a lot of redundancy. Altering this block further may result in lower faithfulness scores for generation tasks.

In conclusion, the selection of a PEFT technique depends on the complexity of the task at hand. For instance, if the task requires reasoning over context \cite{9546640}, it is advisable to choose a method that has attention modules as a workspace. Alternatively, if the task involves the addition of new concepts to the language model, feed-forward modules can be used to store knowledge in the Transformer \cite{dai-etal-2022-knowledge}, thus making them potential workspaces for adaptation. For simple tasks that do not require any of the above requirements, adding task-specific information via the embedding workspace should suffice. All of these insights can be easily deduced using PEFT-Ref.

\subsection{Further development of PEFT techniques}\label{ssec:tech-devel}

Parameter sharing/tying has several advantages in regularisation and inductive biases \cite{pmlr-v151-yeh22b}. ALBERT \cite{Lan2020ALBERT:}, a language model that achieves parameter reduction by sharing and factorising parameters, achieves high performance and stability with fewer parameters than BERT. Consequently, parameter sharing is an attractive property that can significantly contribute to the performance and stability of finetuning techniques. Enabling parameter sharing/tying across layers of different modules, as well as across Transformer layers, holds the potential of significantly enhancing the performance and stability of PEFT techniques.

Adopting a tunable scaling parameter in Adapters, as in LoRa \cite{DBLP:journals/corr/abs-2106-09685}, could dramatically improve these methods as they collaborate with all blocks in the Transformer layer. Such significant collaboration may need to be controlled via scaled addition. We also note simple but effective tweaks, like AdapterDrop \cite{ruckle-etal-2021-adapterdrop}, which dynamically removes some of the Adapter layers that are attached to all Transformer layers in the vanilla settings. Additionally, stability could be increased in prompt tuning by introducing proper layering to produce prompt weights to concatenate with the embeddings.

Another potential direction for development is controlling the number of insertions of a PEFT module by choosing specific layers (rather than all) for insertion. Heuristic specification finetuning techniques (e.g.\ \citeauthor{lee2019elsa}, \citeyear{zaken2022bitfit}, finetune the last quarter of the layers in BERT and RoBERTa) that achieve good performance could be used as indications of which layers to choose. % guide the identification of potential layers where the PEFT techniques should be inserted.

PEFT modules could potentially use the residual flow (i.e.\ contextualised embeddings of the input sequence) as a workspace, and adapt it by either reparameterising or adding a new set of parameters such as scaling vectors. 

In addition, heuristic specification finetuning techniques like BitFit \cite{zaken2022bitfit} and LN-Tuning \cite{qi2022parameterefficient} which finetune the bias terms and LayerNorm in the model respectively, represent potential workspaces for designing PEFT modules to adapt them. The advantage of using PEFT on these heuristic specifications is that it preserves the PLM model's knowledge of parameters like bias and LayerNorm and collaborates with them rather than changing them.

\section{Related Work}\label{section:related_work}
%%% 7. related work and critical comparison with He et al., 2022, and Pfeiffer et al., 2021, etc.

\citet{he2022towards} include a treatment of PEFT methods addressing  internal architecture, modified representations, insertion form, and composition function. However, to fully grasp the potential of PEFT techniques from a modular perspective, embracing a diverse range of properties that compensate for their subtle variations is essential:
%was limited in coverage: 
in the functional form,  all four considered PEFT methods were treated as having the form  Project down → Nonlinear/linear → Project up, but not all PEFT methods have this form (e.g.\ Prompt Tuning, (IA)\textsuperscript{3}, Tiny-Attention Adapters). Moreover, in terms of modified representations, the treatment confusingly treats a Transformer module that produces a hidden representation as a hidden representation in itself (i.e.\ it treats a position (Transformer module) as a hidden representation).  

Additionally, not all PEFT methods modify a hidden representation. In our work, we make a clear separation between the position (the Workspace in PEFT-Ref), and the modified hidden representations (input type in PEFT-Ref). Also not all PEFT techniques are typically integrated with the language model solely through addition forms (e.g.\ Prompt Tuning, (IA)\textsuperscript{3}). %Thus introducing PEFT-Ref, which overcomes prior limitations and includes new characteristics.

\citet{pfeiffer2023modular} present a unified view of modular deep learning, focusing on four key dimensions: module implementation, routing functions, module aggregation, and module training. This perspective revealed connections between previously independent research threads and various applications of modular networks. While \citeauthor{pfeiffer2023modular} briefly discuss some PEFT techniques under module implementation, they only use composition type to categorise them (input composition for prompt and prefix tuning, parameter composition for LoRA, function composition for Adapters). %However, these techniques require more than one characteristic to fully detail them for future design decisions and practical use cases. To address this gap, PEFT-Ref provides a comprehensive view that can serve PEFT techniques and fit easily within the unified view of modular deep learning.

Other work has surveyed parameter-efficient techniques and studied their theoretical underpinnings and performance on various downstream tasks. For example, \citet{Ding2023} design a library on top of the Transformers library \cite{wolf-etal-2020-Transformers} to enable flexible training, composing, attaching/detaching of PEFT techniques with PLMs. \citet{mao-etal-2022-unipelt} propose a mixture of experts framework for PEFT techniques that learn to activate a PEFT technique that best suit the task.

\section{Conclusion}\label{sec:conclusions}

In the work reported here, we aim to contribute to a more comprehensive understanding of the rapidly evolving research area of PEFT techniques.  In this paper, we have introduced the PEFT-Ref framework consisting of a reference architecture and typology based on an inventory of standardised structural and functional properties of PEFT methods. We have shown how PEFT techniques can be characterised in terms of the framework and how such characterisation enables direct comparisons between PEFT methods in terms of efficiency improvements and task performance. 

We further analysed our PEFT-Ref characterisations of seven leading PEFT methods, to (i) draw important conclusions about their suitability for different task types, and (ii) extract clear pointers for developing improved PEFT methods in the future.

%We have proposed the PEFT-Ref framework to support comparisons between PEFT techniques in terms of efficiency and task performance, selection of techniques for downstream tasks, and guidance for design choices in PEFT technique development. %The framework also enables fast comparison of different efficiency aspects and performance of the techniques.
PEFT-Ref provides a simple but general reference architecture desgined to facilitate (i) easy recall of its components, and (ii) comparative understanding of different PEFT methods. Moreover, taking a modular view of PEFT techniques encourages increased reusability of PLMs for various use cases and tasks, and aligns with a recent call to build and maintain large language models like open-source software.\footnote{\href{https://colinraffel.com/blog/a-call-to-build-models-like-we-build-open-source-software.html}{https://colinraffel.com/blog/a-call-to-build-models-like-we-build-open-source-software.html}}

\section*{Limitations}
In this work, our aim is to establish a solid foundation for comprehending PEFT techniques by emphasising a modular view of the parameters they add and/or manipulate. We propose that PEFT techniques can be seen as small modules working in collaboration with large modules, such as language models, to address specific tasks. By adopting this modular perspective, we can capitalise on the structural and functional benefits of modularity.

Our main objective is to unify PEFT techniques, delving deeper into their inner workings to gain a comprehensive understanding. Additionally, we seek to identify areas where these techniques can be improved and offer guidance on making informed choices when selecting a technique for a downstream task.

Regarding pointers for future development we do not (yet) provide implementations for the improvements to PEFT methods we suggest. Regarding relative strengths of different PEFT methods, there are other factors that play into method selection which are 
%b nor can we guarantee 100\% accuracy in the selection process. Such detailed implementation and guarantees are 
beyond the scope of the present work.

Finally, while we have included the leading PEFT methods in our sample characterisations, we have not included all variants and other methods that exist. It is therefore conceivable that their inclusion would lead to modification of the framework, in particular in terms of property value ranges.

%%% 8. discussion including use cases and limitations 
%For future work, we plan to base our framework on the functional properties of modularity such as measuring compositionality, portability, and functionality encapsulation. We aim to increase the reusability of these techniques by guiding the design space of new techniques to account for these functional properties, which could result in significant savings in storage, and training costs.

%\section*{Acknowledgements}

\bibliography{anthology,custom}
\bibliographystyle{acl_natbib}

\appendix
\section{Module Complexity: Further Analysis}
\label{sec:complexity_analysis}
In Table \ref{tab:peft-efficient} and Section \ref{ssec:complexity}, we initially analysed the module complexity of PEFT from the perspective of the time it takes for them to produce their output (forward pass). We also discussed the impact of PEFT-Ref's structural properties on this complexity. In this section, we further extend the analysis and examine how different modules can affect also the backward pass during training, as well as the inference of the language model:

\textbf{Training (forward pass):}
As previously mentioned, the complexity in Table \ref{tab:peft-efficient} represents the number of steps required to generate the PEFT module's output per token. Prompt Tuning and (IA)\textsuperscript{3} only require one step for initialising token-like embeddings weights and scaling vectors, respectively, without additional layering. Therefore, these processes can be disregarded. However, for other techniques, layering is involved from the initialised input to the output, with complexity per token as provided in Table \ref{tab:peft-efficient}. Additionally, it is worth noting that while the attention complexity is $\mathcal{O}(Td)$ per token \cite{NIPS2017_3f5ee243}, Tiny-Attention Adapters use vectors for query, keys, and matrices, resulting in a per-token complexity of $\mathcal{O}(T)$ with $d=1$.

\textbf{Training (backward pass):} For most PEFT modules, it's unnecessary to backpropagate through the entire language model. Whether backpropagation is required or not depends on the location of the PEFT technique's workspace within the language model hierarchy. If the workspace is situated deeper-from backward pass perspective- in the hierarchy, such as in the embedding layer, backpropagation needs to occur at that specific level. For instance, techniques like Prompt Tuning and Prefix Tuning treat the embedding layer as a workspace and necessitate backpropagation to that level.

\textbf{Inference:}
When it comes to inference, Prompt Tuning, Prefix Tuning, and LoRA do not significantly impact latency because we can conceal them behind the inherent latency of the language model. Prompt Tuning and Prefix Tuning techniques require allocation from the model's context window, which falls within the expected processing latency for the size of the context window. In the case of LoRA, as it involves reparameterisation of weights and parallel insertion, we can explicitly compute and store the weights along with their reparameterised version ($W = W_0 + BA$) to facilitate inference as usual \cite{DBLP:journals/corr/abs-2106-09685} .

As for (IA)\textsuperscript{3}, this technique introduces minimal latency since it involves scaling vectors, which are computationally straightforward.

However, Adapters, along with their variants Compacters and Tiny-Attention Adapters, inserted sequentially and processing hidden representations, contribute more substantial latency to the language model compared to other PEFT techniques. To address these implications, \citet{ruckle-etal-2021-adapterdrop} discussed strategies like AdapterDrop and AdapterFusion that can be employed to mitigate the additional latency.
\section{Performance Table}
\label{sec:performance}
\begin{table*}
\centering
\begin{adjustbox}{width=1.2\textwidth,center=\textwidth}
\begin{tabular}{llllcccccc}
\toprule
\textbf{Work}                                                     & \textbf{Task}                        & \textbf{Datasets} &      \textbf{Training Details}                                                                                                                                                                                                                                                                                        & \textbf{LoRA}        & \textbf{\begin{tabular}[c]{@{}c@{}}Prefix \\ Tuning\end{tabular}} & \textbf{\begin{tabular}[c]{@{}c@{}}Prompt \\ Tuning\end{tabular}} & \textbf{(IA)\textsuperscript{3}} & \textbf{Adapter}     & \textbf{Compacter}  \\ \midrule
                                                                  &                                      &                                                                                                                                                                                                                                                                                                                &                      &                                                                   &                                                                   &                                                   &                      &                      &                                                                            \\
\cite{Ding2023} & Sentiment Analysis                   & \begin{tabular}[l]{@{}l@{}}GLUE-SST2\\ ROTTEN\_TOMATOES\\ FINANCIAL\_PHRASEBANK\\ POEM\_SENTIMENT \\ YELP\_POLARITY\end{tabular}                                                                        &  \begin{tabular}[l]{@{}l@{}}Model: T5 base\\ Training Steps: 20k \\except Prompt Tuning trained with 100K steps\\ The best Result from the combination of\\ \{16,32\} batch size, learning rate \{1e-3,1e-4,5e-4\} is taken\end{tabular}                                                                                                          & 93.09                    & 92.83                                                                 & 85.48                                                                 & -                                                 & 92.06                    & -                                                                                              \\ \midrule
                                                                  &                                      &                                                                                                                                                                                                                                                                                                                &                      &                                                                   &                                                                   &                                                   &                      &                      &                                                                            \\
\cite{Ding2023} & Classification/emotion               & \begin{tabular}[l]{@{}l@{}}EMO \\ EMOTION \\ TWEET\_EVAL-HATE \\ TWEET\_EVAL-IRONY\\  TWEET\_EVAL-OFFENSIVE\\  TWEET\_EVAL-SENTIMENT \\ TWEET\_EVAL-STANCE\_ABORTION \\ TWEET\_EVAL-STANCE\_ATHEISM\\  TWEET\_EVAL-STANCE\_CLIMATE \\ TWEET\_EVAL-STANCE\_FEMINIST \\ TWEET\_EVAL-STANCE\_HILLARY\end{tabular}
&  \begin{tabular}[l]{@{}l@{}}Model: T5 base\\ Training Steps: 20k\\except Prompt Tuning trained with 100K steps\\ The best Result from the combination of\\ \{16,32\} batch size, learning rate \{1e-3,1e-4,5e-4\} is taken\end{tabular}  
& 68.70                    & 67.21                                                                 & 52.95                                                                & -                                                 & 68.31                    & -                                                                                             \\ \midrule
                                                                  &                                      &                                                                                                                                                                                                                                                                                                                &                      &                                                                   &                                                                   &                                                   &                      &                      &                                                                            \\
\cite{Ding2023} & Natural Language Inference           & \begin{tabular}[l]{@{}l@{}}ANLI \\ GLUE-MNLI\\  GLUE-QNLI\\  GLUE-RTE\\  SCITAIL\\  SUPERGLUE-RTE\\  SICK \\  SUPERGLUE-CB\end{tabular}   
&  \begin{tabular}[l]{@{}l@{}}Model: T5 base\\ Training Steps: 20k\\except Prompt Tuning trained with 100K steps\\ The best Result from the combination of\\ \{16,32\} batch size, learning rate \{1e-3,1e-4,5e-4\} is taken\end{tabular}  
& 82.73                    & 80.07                                                                 & 51.93                                                                 & -                                                 & 83.06                    & -                                                                                             \\ \midrule
                                                                  &                                      &                                                                                                                                                                                                                                                                                                                &                      &                                                                   &                                                                   &                                                   &                      &                      &                                                                            \\
\cite{Ding2023} & Multiple-Choice QA                   & \begin{tabular}[l]{@{}l@{}}COSMOS\_QA\\  DREAM \\ HELLASWAG \\ OPENBOOKQA \\ QASC \\ QUAREL \\ QUARTZ-NO\_KNOWLEDGE \\ QUARTZ-WITH\_KNOWLEDGE \\ RACE-HIGH \\ RACE-MIDDLE \\ SUPERGLUE-COPA \\ WINO\_GRANDE \\ COMMONSENSE\_QA \\ SCIQ \\ WIQA\end{tabular} 
&  \begin{tabular}[l]{@{}l@{}}Model: T5 base\\ Training Steps: 20k\\except Prompt Tuning trained with 100K steps\\ The best Result from the combination of\\ \{16,32\} batch size, learning rate \{1e-3,1e-4,5e-4\} is taken\end{tabular}  
& 58.67                    & 53.93                                                                 & 46.93                                                                 & -                                                 & 56.11                    & -                                                                                             \\ \midrule
                                                                  &                                      &                                                                                                                                                                                                                                                                                                                &                      &                                                                   &                                                                   &                                                   &                      &                      &                                                                            \\
\cite{Ding2023} & Summarisation                        & \begin{tabular}[l]{@{}l@{}}SAMSUM\\ XSUM\end{tabular}         &  \begin{tabular}[l]{@{}l@{}}Model: T5 base\\ Training Steps: 20k\\except Prompt Tuning trained with 100K steps\\ The best Result from the combination of\\ \{16,32\} batch size, learning rate \{1e-3,1e-4,5e-4\} is taken\end{tabular}                                                                                                                                                                                                                                                   & 35.44                    & 33.61                                                                 & 30.35                                                                 & -                                                 & 35.38                    & -                                                                                             \\ \midrule
                                                                  &                                      &                                                                                                                                                                                                                                                                                                                &                      &                                                                   &                                                                   &                                                   &                      &                      &                                                                            \\
\cite{liu2022fewshot}                            & Commonsense                          & \begin{tabular}[l]{@{}l@{}}H-Swag\\ COPA\\ StoryCloze\\ Winogrande\end{tabular}                                                                                                         & \begin{tabular}[l]{@{}l@{}}Model: T0-3b\\ Training Steps: 1k\\except Compacter/Adapter trained with 500 steps\\ The best Result from the combination of\\ \{8\} batch size, learning rate \{3e-3\} except for Prompt Tuning \{1e-3\}\end{tabular}                                                                                                                       & 66.23                    & 58.95                                                                 & 61.05                                                                 & 68.01                                                 & 63.75                    & 65.65                                                                                              \\ \midrule
                                                                  &                                      &                                                                                                                                                                                                                                                                                                                &                      &                                                                   &                                                                   &                                                   &                      &                      &                                                                            \\
\cite{liu2022fewshot}                           & Word Sense Disambiguation            & WiC                             & \begin{tabular}[l]{@{}l@{}}Model: T0-3b\\ Training Steps: 1k\\except Compacter/Adapter trained with 500 steps\\ The best Result from the combination of\\ \{8\} batch size, learning rate \{3e-3\} except for Prompt Tuning \{1e-3\}\end{tabular}                                                                                                                                                                                                & 54.86                    & 52.51                                                                 & 52.51                                                                 & 54.23                                                 & 54.70                    & 55.33                                                                                             \\ \midrule
\multicolumn{1}{l}{}                                              & \multicolumn{1}{l}{}                 & \multicolumn{1}{l}{}                                                                                                                                                                                                                                                                                           & \multicolumn{1}{l}{} & \multicolumn{1}{l}{}                                              & \multicolumn{1}{l}{}                                              & \multicolumn{1}{l}{}                              & \multicolumn{1}{l}{} & \multicolumn{1}{l}{} & \multicolumn{1}{l}{}                                                       \\
\cite{Ding2023} & Classification/hate-speech detection & \begin{tabular}[l]{@{}l@{}}THOS-DISABILITY \\ ETHOS-GENDER \\ ETHOS-NATIONAL\_ORIGIN \\ ETHOS-RACE \\ ETHOS-RELIGION \\ ETHOS-DIRECTED\_VS\_GENERALIZED \\ HATE\_SPEECH\_OFFENSIVE \\ HATE\_SPEECH\\  HATEXPLAIN\end{tabular}     
&  \begin{tabular}[l]{@{}l@{}}Model: T5 base\\ Training Steps: 20k\\except Prompt Tuning trained with 100K steps\\ The best Result from the combination of\\ \{16,32\} batch size, learning rate \{1e-3,1e-4,5e-4\} is taken\end{tabular}  
& 85.22                    & 84.37                                                                 & 67.69                                                                 & -                                                 & 86.02                    & -                                                                                            \\ \bottomrule
\end{tabular}

\end{adjustbox}
\caption{Average PEFT techniques accuracies on different tasks across datasets. Tiny-Attention Adapters are omitted from the Table due to the absence of comparative studies in the published literature.}
\label{tab:peft-performance}
\end{table*}
\end{document}